%% file: main.tex
\pdfoutput=1

\documentclass[11pt]{article}
\usepackage[table,xcdraw]{xcolor}
\usepackage[]{ACL2023}

\usepackage{times}
\usepackage{latexsym}
\usepackage{algorithm}
\usepackage{algpseudocode}
\usepackage{amsmath}
\usepackage{enumitem}
\usepackage[T1]{fontenc}

\usepackage[utf8]{inputenc}

\usepackage{microtype}

\usepackage{inconsolata}
\usepackage{graphicx}
\usepackage{booktabs}
\usepackage{amsmath}
\usepackage{xcolor}
\usepackage{pdfpages}
\usepackage{afterpage}
\usepackage{stfloats}
\usepackage{xcolor}
\usepackage{amsmath}

\usepackage[normalem]{ulem}
\useunder{\uline}{\ul}{}

\newcommand{\ModelName}{\textsc{Reffly}}

\title{\ModelName{}: Melody-Constrained Lyrics Editing Model} 

\author{First Author \\
  Affiliation / Address line 1 \\
  Affiliation / Address line 2 \\
  Affiliation / Address line 3 \\
  \texttt{email@domain} \\\And
  Second Author \\
  Affiliation / Address line 1 \\
  Affiliation / Address line 2 \\
  Affiliation / Address line 3 \\
  \texttt{email@domain} \\}

\begin{document}
\maketitle
\def\thefootnote{*}\footnotetext{Equal contribution}\def\thefootnote{\arabic{footnote}}
\begin{abstract}

Automatic melody-to-lyric (M2L) generation aims to create lyrics that align with a given melody. While most previous approaches generate lyrics from scratch, \textbf{\textit{revision}}—editing plain text draft to fit it into the melody—offers a much more flexible and practical alternative. This enables broad applications, such as generating lyrics from flexible inputs (keywords, themes, or full text that needs refining to be singable), song translation (preserving meaning across languages while keeping the melody intact), or style transfer (adapting lyrics to different genres).
This paper introduces \ModelName{} (\textbf{RE}vision \textbf{F}ramework \textbf{F}or \textbf{LY}rics), the first revision framework for editing and generating \textbf{melody-aligned} lyrics. We train the lyric revision module using our curated synthesized melody-aligned lyrics dataset, enabling it to transform plain text into lyrics that align with a given melody. To further enhance the revision ability, we propose training-free heuristics aimed at preserving both semantic meaning and musical consistency throughout the editing process. Experimental results demonstrate the effectiveness of \ModelName{} across various tasks (e.g. lyrics generation, song translation), showing that our model outperforms strong baselines, including Lyra \cite{tian2023unsupervised} and GPT-4, by 25\% in both musicality and text quality.

\end{abstract}

\input{content/01-introduction}

\input{content/02-background}
\input{content/03-method}
\input{content/04-experiments_setup}

\input{content/05-results}

\input{content/06-related_works}
\input{content/07-conclusion}

\bibliography{anthology,custom}
\bibliographystyle{acl_natbib}

\appendix
\input{content/Appendix}
\label{sec:appendix}

\end{document}

%% file: content/01-introduction.tex
\section{Introduction}
\begin{figure}[!t]
  \centering
  \includegraphics[width=\columnwidth]{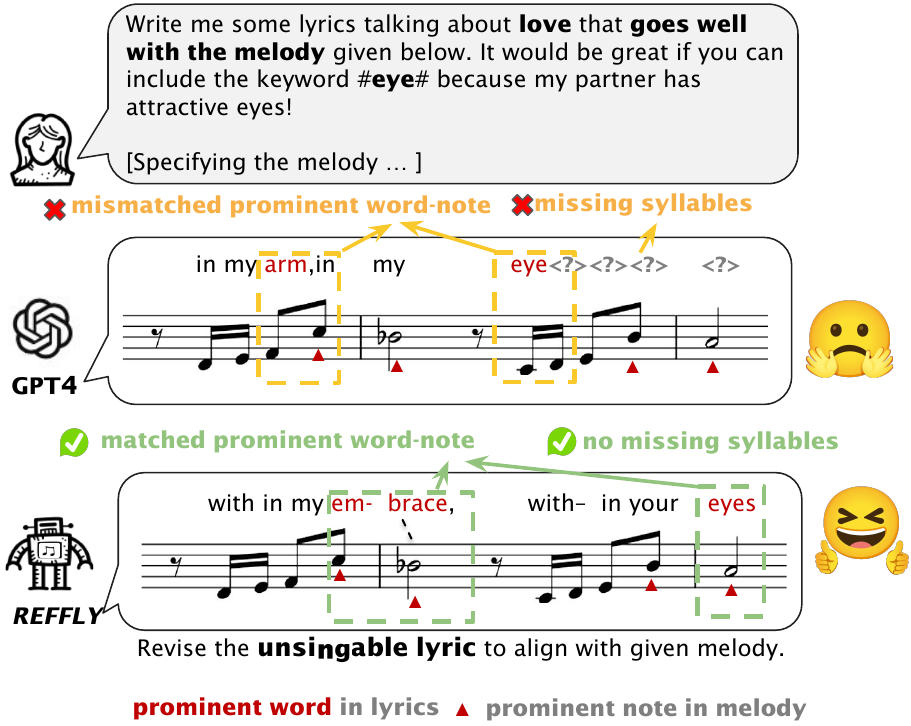}
  \vspace{-8mm}
  \caption{
Human singers naturally emphasize certain words when singing, which align with prominent notes to ensure musical flow (details in $\S$ \ref{sec:def_prominent_note_word}).  
However, LLMs like GPT-4 often misalign these prominent words (e.g."arm" with non-prominent notes) or omit syllables (e.g. no words for the last four notes), lowering lyric quality($\S$ \ref{sec:background}). Our model, \ModelName{}, refines less singable drafts into melodically-aligned lyrics while preserving the meaning. Listen to the \href{https://bit.ly/4fGKWT3}{audios} for an intuitive sense.}
\label{fig:teaser}
\vspace{-5mm}
\end{figure}


Music acts as an important universal language that facilitates social connection and strengthens community involvement \cite{cross2009evolutionary}. Automatic melody-to-lyric (M2L), creating lyrics that are aligned with a given melody, has emerged as a promising task and received interest by the AI community, because it makes the process of music creation more accessible to a wider audience.\citep{sheng2021songmass, tian2023unsupervised, ding2024songcomposer}.

In practice, amateur songwriters may 
wish to craft lyrics that goes with their favorite melody with desired content~\cite{tian2023unsupervised,qian2023unify}, or 
translate songs into different languages for a wider audience or adapt existing lyrics to fit a different melody \cite{Ou2023Songs,nikolov2020rapformer}.
However, existing M2L approaches and AI-assisted songwriting frameworks fall short to support these use cases, due to insufficient control over sentence-level semantic. Most prior works generate lyrics with  zero 
or limited user input such as keywords or topics \citep{sheng2021songmass, tian2023unsupervised, ding2024songcomposer}. Some prior works rely on in-filling text templates without providing sufficient automacy to the user \citep{Zhang2020Youling, Liu2020ChipSong}. In contrast, our revision framework refining plain text into melody-aligned lyrics offers greater flexibility and control. 

In addition, both state-of-the-art LLMs and prior works on lyrics generation struggle with producing singable lyrics that align well with a specific melody. For instance, as illustrated in Figure \ref{fig:teaser}, ChatGPT-4 \cite{openaiGPT-4} generates coherent lyrics but fails to synchronize syllables with the last four music notes. Moreover, prominent words like `arm' and `eye' are paired with less prominent notes, disrupting the overall musical flow and resulting in low prosody. Similarly, prior works on lyrics generation generation either don't consider melody as a constraint~\cite{zhang-etal-2022-qiuniu, 10.1145/3472307.3484175,Zhang2020Youling,Liu2020ChipSong, sun2022songrewriter}, or overlooked the important relationship between prominent note in melody and prominent lyric words \citep{sheng2021songmass,tian2023unsupervised,qian2023unify}, lowering the generated lyric quality. 


Addressing these challenges, we propose a novel \textit{\textbf{revision framework}}, \ModelName{}, which transfers a draft prose to structured and singable lyrics align with a piece of melody. To enhance melody-lyric alignment, we develop a training-free heuristic for capturing prominent lyrical words and musical notes ($\S$\ref{sec:prominent_note}). Since the melody-aligned lyric data is scarce due to copyright constraints, we design an instruction-based mechanism to guide LLMs towards highly singable lyrics by training on a synthetic dataset ($\S$ \ref{sec:training_dataset}). \ModelName{} can generate full-length songs with lyrical verses that develop the song's plot and message, and choruses that repeat a memorable musical motif.

Our contributions are summarized as follows:
\begin{itemize}[topsep=0pt, itemsep=-4pt, leftmargin=*]
    \item  We propose the first melody-constrained lyric revision framework that, given a predefined melody, transfers an arbitrary text (also referred to as a \textit{\textbf{draft}} or \textit{\textbf{unsingable lyrics}}) to a full-length, melody-aligned lyrics with high singability and prosody (also referred to as \textit{\textbf{revised}} or \textit{\textbf{singable lyrics}}), with sentence-level semantic control.
    \item We introduce a training-free heuristic for capturing melody-lyrics alignment, semantically and musically, to improve both \textbf{\textit{singability}} and \textbf{\textit{prosody}}.
    Correspondingly, we also contribute a expert labeled dataset with fine-grained annotations of music sheets. \footnote{Dataset source: \url{https://bit.ly/3X6nCquhttps://bit.ly/3RytMfR}}
    \item In comprehensive experiments 
    across two settings: \textbf{1)} generation of lyrics from user-specified inputs, and \textbf{2)} translation of lyrics from Chinese to English, 
    \ModelName{} 
    significantly enhances lyrics-melody alignment and text quality of the generated lyrics, resulting in a 25\% and 34\% improvements over strong baselines in terms of musicality and overall preference, respectively.\footnote{Demo: \url{https://bit.ly/4fGKWT3}}
\end{itemize}

%% file: content/02-background.tex
\vspace{-0.08in}
\section{Problem Setup and Background}\label{sec:background}
\vspace{-0.03in}
\subsection{What Makes a Good Lyric?}\label{subsec:good_lyrics}
Great lyrics harmonize with the melody, blending musicality (\textit{e.g.,} singability, prosody) with textual quality (\textit{e.g.,} coherence, creativeness) \cite{perricone2018great}. Here, we elaborate the two terms related to musicality below:

\begin{itemize}[topsep=0pt, itemsep=-2pt, leftmargin=*]
    \item \textbf{Singability} is what makes a song easier to sing. For example, it is considered \textit{not} singable when one single music note maps to a multi-syllable word (\textit{e.g.,} beau-ti-ful) in the lyrics \cite{tian2023unsupervised}.
    \item \textbf{Prosody} measures whether melody and lyrics rise and/or fall together \cite{perricone2018great}. Lyrics with good prosody highlights prominent words by matching them with prominent notes. For example, in Figure \ref{fig:teaser}, \ModelName{} enhances expression by stressing prominent words like `embrace' and `eye' by aligning them with prominent notes.
\end{itemize}
These concepts guided the development of heuristics to better align lyrics with the melody.($\S$ \ref{sec:prominent_note}).

\vspace{-0.05in}
\subsection{Task Formulation}
\paragraph{Goal} Given a predefined melody and a plain-text draft, our goal is to revise the unsingable draft into \textit{full-length} lyrics that excel in both musicality and textual quality. 

\vspace{-0.08in}
\paragraph{Formulation} 
We consider full-length songs with the \textit{\textbf{verse-chorus}} structure
For example, the music in Figure \ref{fig:song_strucutre} has the structure of \textit{<verse 1, chorus 1, verse 2, chorus 2>}. Formally, the input melody $\mathbf{M}$ can be defined as a sequence of $T$ substructures $\mathbf{M}=\{\mathcal{M}_{<tag_1>}, ..., \mathcal{M}_{<tag_T>}\}$, $tag_i \in \{\text{verse, chorus}\}$.  Each $\mathcal{M}_{<tag_i>}$ consists of $K_i$ music phrases (\textit{i.e.,} $\mathcal{M}_{<tag_i>}=\{ p_{i1}, p_{i2}, ..., p_{iK_i}\}$), where each music phrase further contains $N_{ij}$ music notes (\textit{i.e.,} $p_{ij} = \{n_{ij_1}, n_{ij_2}, ... n_{ij_{N_{ij}}}\}$). Here, each music note has three attributes: pitch (\textit{i.e.,} how high or low it sounds), duration (\textit{i.e.,} how long it lasts), and offset (\textit{i.e.,} when it starts). 
The output is lyrics $\mathcal{L}$ that aligns with the input melody at the all granularities (\textit{i.e.,} music notes, phrases, and substructures): $\mathcal{L} = \{ w_{11_1}, w_{11_2}, ...,w_{ij_l}, ..., w_{T{K_T}_N}\}$. Here, $ w_{ij_l}$ is a word or a syllable of a word that aligns with the music note $n_{ij_l}$.

\vspace{-0.08in}

%% file: content/03-method.tex
\begin{figure*}[t]
\centering
\includegraphics[width=\textwidth]{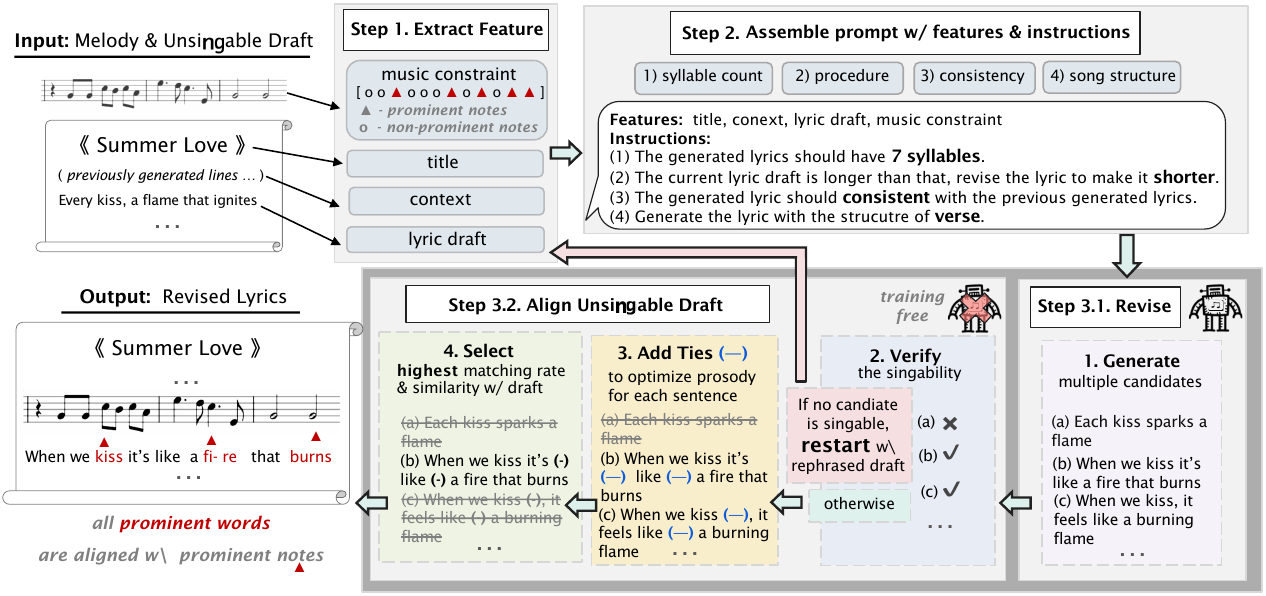}\vspace{-2mm}
\caption{The overview of the inference process of \ModelName{}, an iterative approach to revise each sentence from the unsingable draft based on corresponding music constraint that is extracted from the music score. \ModelName{} begins by taking the melody and the unsingable draft as inputs. It then extracts features and constructs a prompt (Steps 1 and 2). Subsequently, it prompts a trained revision module to revise the unsingable draft (Step 3.1) and aligns the revised draft with the melody constraints using an alignment algorithm (Step 3.2). \textbf{\textit{Note that only Step 3.1 requires training, and all other processes are training-free}}.} 
\label{fig:overview}\vspace{-5mm}
\end{figure*}

\section{Revision Framework for Lyrics}\label{sec:method} 


Figure \ref{fig:overview} illustrates the inference process of \ModelName{}. To manage the complexity of lyric revision, the revision process is conducted at the sentence level. We iteratively revise each sentence from the unsingable draft to lyric that fits the melody, aligning the prominent words with prominent notes while maintaining the overall coherence.


In this section, we detail each component of \ModelName{}. We develop a training-free heuristic for capturing prominent lyrical words and musical notes (Section $\S$\ref{sec:prominent_note}). Then, $\S$\ref{sec:revision_model} introduces our lyrics revision module that refines unsingable drafts based on musical constraints. Last, $\S$\ref{sec:alignement} provides an overview of the inference process to achieve optimal lyric-melody alignment. 

\vspace{-0.03in}
\subsection{Aligning Melody with Lyrics}
\label{sec:prominent_note}

Building on the way experienced singers emphasize certain lyrics to enhance their connection with the melody for musical expressiveness \cite{Robinson2005}, we develop a heuristic to align prominent words with prominent musical notes.\footnote{It is not feasible to use a neural network-based method for aligning prominent words with prominent musical notes due to the lack of such annotated data.} This subsection outlines the process of identifying these prominent notes and words, for which we constructed an expert-annotated dataset to evaluate their effectiveness (see $\S$\ref{sec:effectiveness_heuristics} for results). These heuristics are then used in lyrics generation ($\S$\ref{sec:alignement}) and the construction of synthetic training data ($\S$\ref{sec:training_dataset}).

\paragraph{Extracting Prominent Musical Notes}
\label{sec:def_prominent_note_word}
We identify prominent musical notes that stand out in melodies based on three fundamental characteristics of music:
\textit{Time signature}, \textit{Rhythm}, and \textit{Pitch} \cite{Palmere4cfret2006what}. A musical note is considered prominent if it appears on a stressed beat location as defined by its time signature, exhibits syncopation, or having a large pitch jump from the proceeding notes (More details in Appendix  \ref{subsec:prominent_note_details}).
\vspace{-0.06in}
\paragraph{Extracting Prominent words from Lyrics}
According to \citet{Reikofski2015SingingIE}, nouns, verbs, adjectives are crucial for effectively conveying meaning. Therefore, we identify nouns, verbs, and adjectives that are non-stop words as prominent words. 
\vspace{-0.06in}
\paragraph{Assessing the Accuracy}
\label{sec:valid_dataset}
To the best of our knowledge, we are the first to apply computational algorithms to identify prominent notes and words. Therefore, to evaluate the accuracy of our heuristic, we \textbf{collected a validation dataset} consisting of 100 song clips, each containing three to five musical phrases, annotated by professional musicians marking all the important notes in each melody. This dataset covers a diverse range of music styles, including Jazz, Country, Blues, Folk, Pop, and Comedy. Figure~\ref{fig:val_datapoint} shows an exemplary data point.

\vspace{-0.03in}
\subsection{Lyrics Revision Module}
\label{sec:revision_model}
\label{sec:song_strucutre}
\begin{figure}[!t]
  \centering
  \includegraphics[width=\columnwidth]{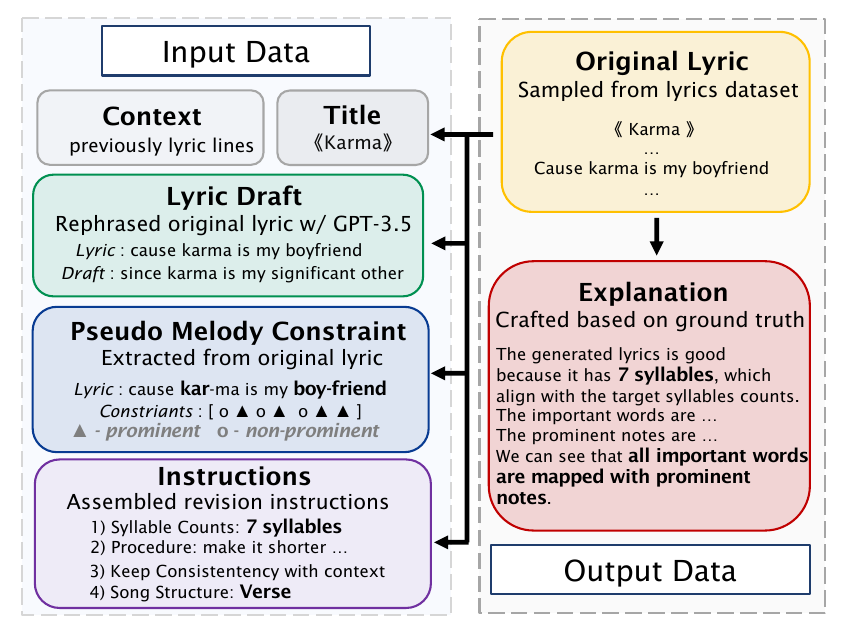} \vspace{-8mm}
  \caption{An exemplary data point in the fine tune dataset. The task is to use a rephrased input, title, music constraint, previously generated lyrics, and assambled instruction to generate the original lyrics, and some explanation. Rephrasing is done by GPT-3.5. During training, the revision model is guided by pseudo melody constraints derived from the original lyrics, enabling it to follow real melody constraints during inference.}
\label{fig:traing} \vspace{-0.18in}
\end{figure}

To achieve high-quality lyrics revision, we fine-tuned a LlaMA2-13b-chat model \cite{touvron2023llama} to effectively transform an unsingable draft into lyrics that fit a given melody while preserving the original meaning. We address three main tasks:

(1) \uline{Syllable planning}: Generating sentences with the necessary syllable count to ensure singability. (2) \uline{Aligning prominent words with notes}: Matching the stressed syllable of prominent words with prominent notes to enhance prosody. (3) \uline{Maintaining local and global coherence}: Ensuring smooth transitions between sentences for local coherence and capturing global structures, such as verse and chorus, for structure-awareness. Addressing these tasks is challenging for LLMs, which are known to struggle with numerical planning \cite{sun2023evaluating}. Furthermore, we face a lack of labeled datasets to train a supervised model that learned to map the prominence in lyrics and melody. 

To empower the model with the first two abilities, we propose a ``\uline{pseudo music constraint}'' (blue box of Figure~\ref{fig:traing}) to improve the syllable planning and word-note matching. The pseudo constraint, derived from generated lyrics, indicates prominent note positions and syllable counts. During training, the model follows pseudo constraints, while in inference, it  applies melody constraints. (refer to $\S$\ref{app:training_data} for more details). 

To maintain both local and global coherence, we introduce an instruction template (purple box of Figure~\ref{fig:traing}). To ensure local, sentence-level, coherency, we provide the model with previously generated lyrics and the song's title as context for each training data point.To enable full-length song generation, our framework incorporates \textbf{structure-awareness} by embedding song structure into the fine-tuning phase. During this phase, we use song structure information ( introduced in section \ref{subsec:good_lyrics}) from the data source for each lyric line, allowing the model to recognize features of different song structures like verses and choruses. Structure tags are embedded within the instruction component and integrated into the prompt.



\subsection{Generate Lyrics at Inference Time}
\label{sec:alignement}



As is illustrated in figure \ref{fig:overview}, \ModelName{} takes the melody and each sentence from unsingable draft as input, to produce singable lyrics as the output, with three steps as following.
\vspace{-0.02in}
\paragraph{Step 1: Extract Features.} Using heuristic in $\S \ref{sec:prominent_note}$ we identify prominent musical notes from the melody, encoding them into a melody constraint. We then prepare this music constraint, title, previously generated lyrics as context, and lyrics draft as input features for further assembling prompt. 
\vspace{-0.02in}
\paragraph{Step 2: Assemble Prompt.} We assemble instructions that specify various features for the desired lyrics, such as matching the syllable count given melody, refining lyrics, providing previous context to ensure coherence, and guidance to maintain desired song structures. An example input prompt can be found in ~Figure \ref{fig:example_training_detail} in Appendix.

\vspace{-0.02in}
\paragraph{Step 3: Revise and Align.}
To enhance the model's ability to generating singable, prominence-aligned lyrics, we break down the process into two sub-steps, iteratively generating the lyrics.

\uline{Step 3.1 Revise}: We adopt diverse beam search \cite{vijayakumar2016diverse} to generate multiple candidate revisions of the unsingable draft, evaluating each for singability. A lyric is singable if: 1) each note corresponds to one or zero syllables; 2) each syllable in multi-syllable words matches a note no longer than a half-note; 3) multi-syllable words do not cross rests. If no candidates meet these criteria, we restart the process with a rephrased draft.

\begin{algorithm}[!t]
\small
\begin{algorithmic}[1]
\State \textbf{input:} List of candidates $C$, orignal draft $o$, melody constraint $m$, max number of ties $K$
\State \textbf{output:} Revised singable lyric
\\
\State $c_{qualified}$ = empty list
\For{candidate $c$ in $C$}
    \State $c_{num}=$ calculate\_num\_ties$(c, m)$ 
    \If{ $ 0 \leq c_{num}  \leq K$}
        \State $c_{tie}= add\_tie (c, m, c_{num})$
        \State $c_{qualified} += c_{tie}$
    \EndIf
\EndFor

\For{candidate $c$ in $c_{qualified}$}
    \State $c_{best}=$argmax(sim($c_{best},o)$, sim$(c,o))$
\EndFor
\State return $c_{best}$
\end{algorithmic}
\caption{Candidate Selection}\vspace{-1mm}
\label{algo:alignment}
\end{algorithm}

\uline{Step 3.2 Align Unsingable Draft}: Algorithm \ref{algo:alignment} 
illustrates the alignment algorithm. For each qualified candidate $c$ and music constraint $m$, we determine the number of ties (a common musical notation that maps more than one notes to one syllable) to add using the following:
\vspace{-2mm}
\begin{equation}
\vspace{-2mm}
\#\text{Ties} = \#\text{Notes}(m) - \#\text{Syllables}(c)
\end{equation}
Next, we define $K$, a tune-able hyper parameter, as maximum number of ties allowed within each musical phrase. We set $K=2$ as a reasonable number in all of our experiments. If $\#\text{ties} < 0$ or $\#\text{ties} > K$, we reject the input. Otherwise, we explore all feasible positions to insert ties, aiming to maximize the number of prominent words mapped to prominent notes.

Finally, we select the candidate whose most important words align with prominent notes. If multiple candidates align perfectly, we choose the one most similar to the original sentence based on BERTScore \cite{zhang2019bertscore}.

%% file: content/04-experiments_setup.tex
\vspace{-2mm}
\section{Experiments setup}\label{sec:experiment}


\subsection{Synthetic Training Dataset}


\label{sec:training_dataset}
As shown in Figure~\ref{fig:traing}, the objective of our training dataset is to instruct the model to generate original lyrics by revising draft lyrics, following music constraints. We construct this dataset using 3,500 song-lyrics collected from the internet. Notably, our revision model only requires lyrics during training, alleviating the lack of aligned melody-lyrics data and potential copy-right issue.\footnote{More details about the different components of input and output of training dataset can be found in Appendix \ref{app:training_data}.}

 
\subsection{Tasks Setup} 
Our model's versatility is demonstrated through its performance across three distinct tasks. We prompt LLaMA2-13b in a few-shot manner to generate lyrics drafts based on user thoughts\footnote{Details on generating the lyrics draft are in Appendix  \ref{subsec:input-to-draft}}:

\textbf{1. Lyrics generation from arbitrary content.}  This task generates song lyrics from scratch, starting with a draft based on scattered user thoughts. The lyrics' quality and melody alignment are evaluated using automated metrics and human judgment.

\textbf{2. Full-Length generation with song structures (Structure-Aware Generation)} In this task, we generate lyrics with specific structural requirements, starting from scattered user feedback. Domain experts then review these generated lyrics for coherence and clarity.

\textbf{3. Song Translation}: This task focuses on translating lyrics from Chinese to English. The initial draft is a straightforward text translation produced by a translation model. We recruit bilingual evaluators to assess the translated lyrics quality.



\vspace{-0.1in}
\subsection{Compared Models}
We compare our framework with two \textbf{baselines} and introduce two \textbf{ablation} \textbf{variations} of \ModelName{} to validate each component.

\textbf{Baselines}.
\textbf{1. Lyra} is an unsupervised, hierarchical melody-conditioned lyric generator that can generate high-quality lyrics with content control without training on melody-lyric data \cite{tian2023unsupervised}.
\textbf{2. SongMass} is an LLM design that leveraging masked sequence to sequence (MASS) pre-training and attention based alignment modeling for lyric-to-melody and melody-to-lyric generation \cite{sheng2021songmass}. 
\textbf{3. GPT-4} is a strong versatile LLM \cite{openaiGPT-4} to compare with. We utilize few-shot prompt to provide a template and instruct the model to follow it.

\textbf{Variations}.
\textbf{4. \ModelName{} w/o S.} is a variant of our proposed framework without the candidate selection algorithm (shown in the green and yellow boxes of Step 3.2 in Figure~\ref{fig:overview}).
\textbf{5. \ModelName{} w/o I.} excludes the instruction component during training (purple and red boxes in Figure~\ref{fig:traing} ).\footnote{We describe details of baselines in Appendix \ref{sec:baselines_detail}.}

\subsection{Evaluation Setup}
We conduct both automatic and human evaluations to assess our framework. While human evaluation is more reliable, it is difficult to scale and reproduce. Therefore, we use widely-adopted metrics like diversity, perplexity, and BERTScore to evaluate creativity, smoothness, and semantic similarity between generated lyrics and initial drafts \citep{sheng2021songmass, tian2023unsupervised}.
\subsubsection{Automatic Evaluation}

We evaluate the generated lyrics on text quality and melody alignment. For text quality, we assess several aspects: 1) \textbf{Diversity}, measured by calculating the number of unique n-grams in the text; 2) \textbf{Perplexity}, 
using GPT-2 to evaluate fluency and predictability; 3) \textbf{Similarity}, 
evaluated with BERTScore, to measure the similarity between our model-generated lyrics and the lyrics draft. For melody alignment, we proposed the \textbf{prominent word-note matching rate}, as explained in $\S$ \ref{subsec:good_lyrics}, to measure how well prominent words are aligned with prominent musical notes.

\begin{table*}[!t]
\small
\centering
\setlength{\tabcolsep}{1.5mm}
\begin{tabular}{@{}l|ccccc|ccccc@{}}
\toprule
\rowcolor[HTML]{EFEFEF} &
  \multicolumn{5}{c|}{\textbf{Automatic Evaluation}} &
  \multicolumn{5}{c}{\textbf{Human Evaluation}} \\ \cline{2-11}
\textbf{Model} &
  \multicolumn{1}{c}{\begin{tabular}[c]{@{}c@{}}Diversity\\ (Unigram)$\uparrow$\end{tabular}} &
  \multicolumn{1}{c}{\begin{tabular}[c]{@{}c@{}}Diversity\\ (Bigram)$\uparrow$\end{tabular}} &
  \multicolumn{1}{c}{\begin{tabular}[c]{@{}c@{}}Similar-\\ ity$\uparrow$\end{tabular}} &
  \multicolumn{1}{c}{\begin{tabular}[c]{@{}c@{}}Perplex-\\ ity$\downarrow$\end{tabular}} &
  \begin{tabular}[c]{@{}c@{}}Match\\ Rate$\uparrow$\end{tabular} &
  \multicolumn{1}{c}{Prosody$\uparrow$} &
  \multicolumn{1}{c}{Coherence$\uparrow$} &
  \multicolumn{1}{c}{\begin{tabular}[c]{@{}c@{}}Intelli-\\ gibility$\uparrow$\end{tabular}} &
  \multicolumn{1}{c}{\begin{tabular}[c]{@{}c@{}}Singab-\\ ility$\uparrow$\end{tabular}} &
  \begin{tabular}[c]{@{}c@{}}Creati-\\ vity$\uparrow$\end{tabular} \\ \midrule
\rowcolor[HTML]{EFEFEF} Lyra &
  \multicolumn{1}{c}{0.52} &
  \multicolumn{1}{c}{0.86} &
  \multicolumn{1}{c}{0.72} &
  \multicolumn{1}{c}{3305} &
  0.48 &
  \multicolumn{1}{c}{1.97} &
  \multicolumn{1}{c}{1.66} &
  \multicolumn{1}{c}{2.02} &
  \multicolumn{1}{c}{1.83} &
  1.70 \\ 
SongMASS &
  \multicolumn{1}{c}{0.50} &
  \multicolumn{1}{c}{0.76} &
  \multicolumn{1}{c}{—} &
  \multicolumn{1}{c}{3759} &
  0.40 &
  \multicolumn{1}{c}{1.35} &
  \multicolumn{1}{c}{1.11} &
  \multicolumn{1}{c}{1.65} &
  \multicolumn{1}{c}{1.46} &
  1.07 \\ 
\rowcolor[HTML]{EFEFEF} GPT-4 &
  \multicolumn{1}{c}{0.51} &
  \multicolumn{1}{c}{0.81} &
  \multicolumn{1}{c}{{\ul 0.83}} &
  \multicolumn{1}{c}{{\ul 635}} &
  0.35 &
  \multicolumn{1}{c}{1.63} &
  \multicolumn{1}{c}{1.96} &
  \multicolumn{1}{c}{1.59} &
  \multicolumn{1}{c}{1.45} &
  1.92 \\ \midrule

\ModelName{} w/o S. &
  \multicolumn{1}{c}{{\ul 0.54}} &
  \multicolumn{1}{c}{\textbf{0.88}} &
  \multicolumn{1}{c}{0.78} &
  \multicolumn{1}{c}{1226} &
  0.51 &
  \multicolumn{1}{c}{{\ul 2.12}} &
  \multicolumn{1}{c}{{\ul 2.27}} &
  \multicolumn{1}{c}{{\ul 2.24}} &
  \multicolumn{1}{c}{{\ul 2.29}} &
  {\ul 2.06} \\ 
\rowcolor[HTML]{EFEFEF} \ModelName{} w/o I. &
  \multicolumn{1}{c}{0.51} &
  \multicolumn{1}{c}{0.81} &
  \multicolumn{1}{c}{0.74} &
  \multicolumn{1}{c}{635} &
  {\ul 0.59} &
  \multicolumn{1}{c}{1.98} &
  \multicolumn{1}{c}{2.01} &
  \multicolumn{1}{c}{1.87} &
  \multicolumn{1}{c}{1.93} &
  1.74 \\ 
\ModelName{} &
  \multicolumn{1}{c}{\textbf{0.59}} &
  \multicolumn{1}{c}{{\ul 0.87}} &
  \multicolumn{1}{c}{\textbf{0.84}} &
  \multicolumn{1}{c}{\textbf{310}} &
  \textbf{0.82} &
  \multicolumn{1}{c}{\textbf{2.27}} &
  \multicolumn{1}{c}{\textbf{2.46}} &
  \multicolumn{1}{c}{\textbf{2.35}} &
  \multicolumn{1}{c}{\textbf{2.32}} &
  \textbf{2.22} \\ 
  \bottomrule
\end{tabular}
\vspace{-2mm}
\caption{Evaluation Results for the Arbitrary Generation task. \ModelName{} and its variants (\ModelName{} w/o S. and \ModelName{} w/o I.) consistently outperform other models across most metrics, both in automatic and human evaluations. }\vspace{-2mm}
\label{tbl:evaluation_result}
\end{table*}
\vspace{-0.09in}
\subsubsection{Human Evaluation}
\paragraph{Annotation Task} We conducted a qualification task to select annotators with expertise in song and lyric annotation on Mechanical Turk. Additional details on the qualification process are provided in Appendix \ref{app:qual}. Our annotation process is comparative, with annotators reviewing groups of songs produced by various systems that share the same melody and title. All baseline models were assessed. At least three workers annotated each song, rating the lyrics' quality on a 1-5 Likert scale across five categories. For musicality, the workers assessed \textbf{prosody} (whether prominent words were exaggerated by melody), \textbf{intelligibility} (whether the lyric content was easy to understand when listen to it),  and \textbf{singability} (how clearly the lyrics could be understood). In terms of text quality, they evaluated \textbf{coherence} and \textbf{creativity}. Annotators evaluated \textbf{structural clarity} (whether the verse-chorus structure is clear) structural-aware generation and assessed \textbf{translation quality} in song translation. The average inter-annotator agreement in terms of Pearson correlation was 0.69.

%% file: content/05-results.tex
\vspace{-0.1in}
\section{Results}

\begin{table}[t!]
\centering
\small
\begin{tabular}{@{}c|c|c@{}}
\toprule
 & \begin{tabular}[c]{@{}c@{}}Note Extraction\\ Success Rate\end{tabular} & \begin{tabular}[c]{@{}c@{}}Alignment \\ Success Rate\end{tabular} \\ \midrule
Duration-Only & 74\% & 43\% \\
Comprehensive w/o adj.  & 96\% & 65\% \\ 
Comprehensive    & \textbf{96\%} & \textbf{91\%} \\ \bottomrule
\end{tabular}
\caption{Comparison of three extraction and alignment strategies. The highest performance in each category is highlighted in bold, illustrating the superior effectiveness of our strategy in both note extraction (96\%) and alignment (91\%).}
\label{tab:lyric-melody alignment}
\vspace{-5mm}
\end{table}
\subsection{Effectiveness of the proposed heuristic}
\label{sec:effectiveness_heuristics}
Our heuristic for identifying prominent notes and words is validated against baselines using a musician-annotated dataset of 100 song clips ($\S$ further details in \ref{sec:valid_dataset}). The first baseline (Duration-only) relies solely on note duration to determine prominence, similar to the decoding constraints used in Lyra \cite{tian2023unsupervised}, and pairs this with our word extraction heuristic. The second baseline (Comprehensive w/o adj.) utilizes our note extraction heuristic but restricts prominent words to nouns and verbs.

As shown in Table \ref{tab:lyric-melody alignment}, our prominent note extraction heuristic achieves an accuracy of 96\%, substantially outperforming both baselines. Furthermore, our comprehensive heuristic for extracting prominent words and notes yields a 91\% alignment
success rate\footnote{Alignment
success rate is the accuracy of prominent words correctly mapped to prominent notes; note extraction success rate is the accuracy of extracting prominent notes}, surpassing the best baseline by 26\%. These results underscore the effectiveness and non-trivial nature of our approach in capturing the alignment between prominent words and prominent notes (more details in Appendix \ref{subsec:Results_heuristics_details}).

\subsection{Result of Lyrics Generation from Arbitrary Content}
\label{sec:automatic_eva}

The results of automatic evaluation (mainly assesses fluency, topic relevance, and melody-lyric alignment) and human evaluations (assesses overall quality across multiple aspects such as musicality, creativity, etc.) are reported in Table~\ref{tbl:evaluation_result}. 

\paragraph{Automatic Results}
The similarity scores in Table~\ref{tbl:evaluation_result} indicate that \ModelName{} and GPT-4 excel in preserving the meaning of unsingable drafts. In contrast, SongMASS and Lyra surpass GPT-4 in terms of musicality, but at the cost of fluency. The qualitative example (shown in Section \ref{sec:case_study}) shows that SongMASS and Lyra tend to generate cropped sentences to fit the music, leading to higher perplexity. Although GPT-4 matches \ModelName{} in retaining lyrical meaning, it falls short in diversity and melody alignment, as reflected in its lowest Match Rate. Overall, \ModelName{} surpasses all baselines, producing lyrics with superior textual quality, optimal melody alignment, and faithful preservation of the original draft's meaning.


\begin{table*}[!t]
\small
\begin{tabular}{@{}ccccccccccc@{}}
\toprule
 &
  \multicolumn{4}{c}{Automatic Evaluation} &
  \multicolumn{6}{c}{Human Evaluation} \\ \midrule
\multicolumn{1}{c|}{Model} &
  \begin{tabular}[c]{@{}c@{}}Diversity$\uparrow$\\ (Unigram)\end{tabular} &
  \begin{tabular}[c]{@{}c@{}}Diversity$\uparrow$\\ (Bigram)\end{tabular} &
  \begin{tabular}[c]{@{}c@{}}Perplex-\\ ity$\downarrow$\end{tabular} &
  \multicolumn{1}{c|}{\begin{tabular}[c]{@{}c@{}}Match\\ Rate$\uparrow$\end{tabular}} &
  Prosody$\uparrow$ &
  \begin{tabular}[c]{@{}c@{}}Coher-\\ ence$\uparrow$\end{tabular} &
  \begin{tabular}[c]{@{}c@{}}Intelli-\\ gibility$\uparrow$\end{tabular} &
  \begin{tabular}[c]{@{}c@{}}Singab-\\ ility$\uparrow$\end{tabular} &
  \begin{tabular}[c]{@{}c@{}}Creati-\\ vity$\uparrow$\end{tabular} &
  \begin{tabular}[c]{@{}c@{}}Translate\\ Quality$\uparrow$\end{tabular} \\ \midrule
\rowcolor[HTML]{EFEFEF} 
\multicolumn{1}{c|}{\cellcolor[HTML]{EFEFEF}GPT-4} &
  0.50 &
  0.76 &
  522 &
  \multicolumn{1}{c|}{\cellcolor[HTML]{EFEFEF}0.35} &
  1.59 &
  2.24 &
  1.83 &
  1.54 &
  2.26 &
  2.33 \\
\multicolumn{1}{c|}{\ModelName{}} &
  \textbf{0.59} &
  \textbf{0.87} &
  \textbf{310} &
  \multicolumn{1}{c|}{\textbf{0.83}} &
  \textbf{3.28} &
  \textbf{3.08} &
  \textbf{3.25} &
  \textbf{3.08} &
  \textbf{2.69} &
  \textbf{3.04} \\ \bottomrule
\end{tabular}
\vspace{-1mm}
\caption{Song translation task result. \ModelName{} scores the highest for all metrics. }\vspace{-5mm}
\label{tbl:translation}
\end{table*}

\begin{table}[!t]
\centering
\small
\begin{tabular}{@{}lcc@{}}
\toprule
                   & \textbf{GPT-4} & \textbf{\ModelName{}}  \\ \midrule
Prosody            & 1.31  & \textbf{3.10} \\
\rowcolor[HTML]{EFEFEF}Sinability         & 1.28  & \textbf{3.11} \\
Coherence          & 1.84  & \textbf{2.70} \\ 
\rowcolor[HTML]{EFEFEF}Creativity         & 1.85  & \textbf{2.62} \\ 
Intelligibility    & 1.26  & \textbf{2.99} \\ 
\rowcolor[HTML]{EFEFEF}Structural Clarity & 1.15  & \textbf{3.36} \\ \bottomrule
\end{tabular}
\vspace{-1mm}
\caption{Structure-aware generation results: \ModelName{} outperforms GPT-4 by producing lyrics with a clearer song structure while maintaining lyric quality and melody alignment.}
\vspace{-6mm}
\label{tbl:strucure}
\end{table}
\vspace{-0.05in}
\paragraph{Human Evaluation Results}
\label{sec:human_eva}
For melody-alignment quality, \ModelName{} achieves the highest scores in prosody, singability, and intelligibility. Lyra performs adequately but falls short compared to 
\ModelName{}, as it does not align prominent words and notes during generation. SongMASS and GPT-4 have much lower scores, suggesting that their lyrics may not fit well with the melody. This indicates that \ModelName{} excels in generating lyrics that align well with the melody are easy to sing.

For text quality, \ModelName{} scores the highest in both creativity and coherence, indicating its ability to generate lyrics that are both creative and contextually consistent. While GPT-4 performs reasonably well in text quality, its musicality remains poor. The other models score low in coherence, suggesting their lyrics may lack logical progression and contextual consistency.


\vspace{-0.05in}
\subsection{Ablation Study}
\label{sec:ablation}
We conduct an ablation study to validate each component in \ModelName{}. A qualitative example in Figure~\ref{fig:example_ablation} (Appendix) compares variations of the model. Candidate selection algorithm select optimal candidate generated from revision model (see Figure~\ref{fig:overview}). Instruction component in training simplifies the task for the revision module and enabling better context awareness. As shown in Table~\ref{tbl:evaluation_result}, removing the candidate selection (\ModelName{} w/o S.) or instruction mechanisms (\ModelName{} w/o I.) results in a noticeable performance decline in almost every metrics compared with \ModelName{}.
\vspace{-0.03in}
\subsection{Result of Structure-Aware Generation}
Among all baselines, only GPT-4 has the capability of generating full-length structured lyrics. Table \ref{tbl:strucure} shows the results for the structure-aware generation task, where \ModelName{} achieved a 44\% improvement in structural clarity. Figure ~\ref{fig:example_struct_output} in Appendix shows the generated lyrics's clear verse-chorus-verse-chorus structure.
\subsection{Result of Song Translation}

Similar to the previous task, only GPT-4 has the capability of song translation, so it is our only baseline.
Table \ref{tbl:translation} presents the evaluation result. \ModelName{} demonstrates a remarkable on average 23\% increase in all human evaluation metrics.
This results suggest \ModelName{} significantly enhances the quality, coherence, and melody-alignment of generated lyrics compared to GPT-4, making it more suitable for practical applications in song translation.

\begin{figure*}[t]
\centering
\includegraphics[width=\textwidth]{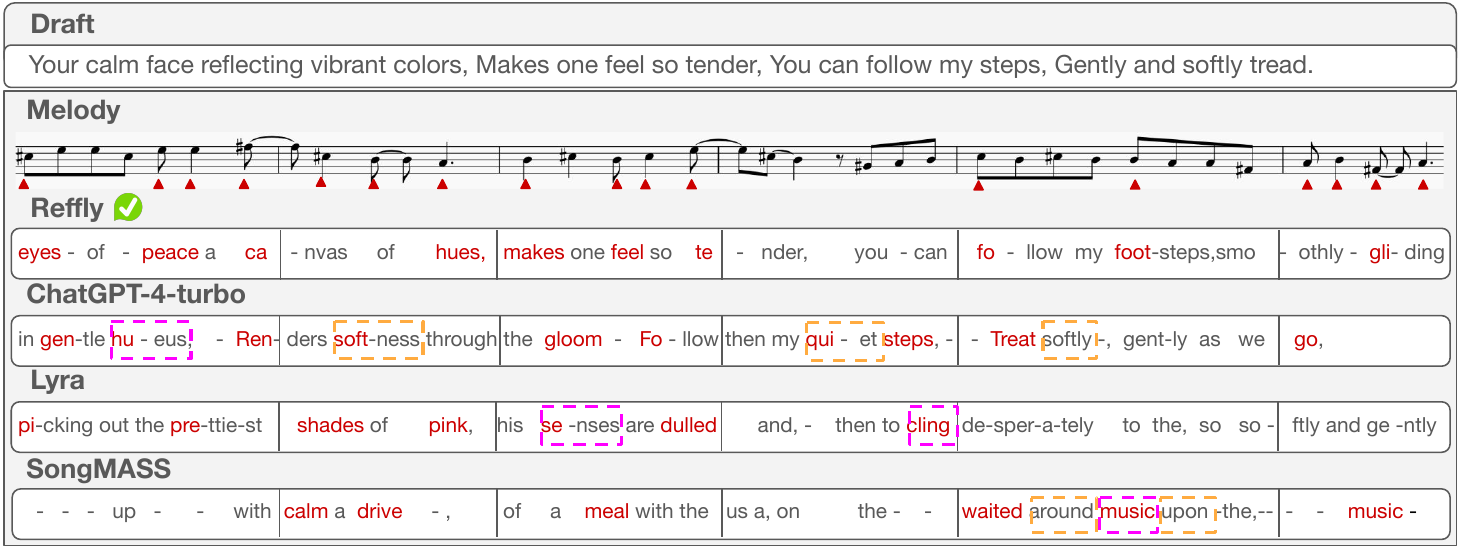}
\vspace{-0.2in}
\caption{The output of different models given the same input draft. \ModelName{} is the only model that aligns lyrics with the melody while preserving the original meaning. Other models produce unsingable or low-prosody lyrics (introduced in ~Section~\ref{subsec:good_lyrics}). The \textcolor{orange}{orange} box highlights the lyrics that is \textbf{not} \textbf{\textit{singable}}. For example, in SongMASS generated lyrics, `a-round', a two-syllable word, is mapped to one musical note, making it hard to sing.  The \textcolor{purple}{purple} box highlights important words that \textbf{failed} to map to a prominent musical note (\textbf{low} \textbf{\textit{prosody}}), which would disrupt the expressiveness of lyrics. Listen to the \href{https://bit.ly/4fGKWT3}{audios in the demo page} for an intuitive sense. 
}
\label{fig:example_case_study}
\vspace{-0.17in}
\end{figure*}

Interestingly, although GPT-4 generally has stronger translation abilities compared to LLaMA2-13b, \ModelName{} outperforms GPT-4 in translation quality by 14\%. This suggests that successful translation in the lyrics-writing context requires not only high text quality, but also an emphasis on how well the lyrics sound when singing, as shown in Figure~\ref{fig:example_case_study}

\vspace{-0.07in}
\section{Case Study}
\vspace{-0.07in}
\label{sec:case_study}
We conducted a case study to better understand the advantages of Reffly compared to baselines. An exemplary generated output is shown in Figure~\ref{fig:example_case_study}.

\noindent{\textbf{Musicality}}: Our model generates melody-aligned lyrics while preserving the original draft's meaning. Unlike the baselines, which overlook the importance of mapping prominent words to prominent notes, our approach ensures that the melody emphasizes these words. In Figure~\ref{fig:example_case_study}, the purple boxes highlight important words that are \textit{failed} to map with prominent notes, and only \ModelName{} generate melody-aligned lyrics. In addition, SongMASS and ChatGPT-4 generate \textit{unsingable} lyrics, indicated by yellow box.

\noindent{\textbf{Revision capability:}} 
 \ModelName{} rephrases sentence structures or modifies words, adding ties to ensure prominent word-note alignment. For example, in Figure~\ref{fig:example_case_study}, it rephrased `Your calm face reflecting vibrant colors' into `eyes of peace, a canvas of hues', and `You can follow my steps' into `You can follow my footsteps'. In both cases, the original meaning is retained.
Although both \ModelName{} and Lyra generate lyrics line-by-line, \ModelName{} produces coherent lyrics due to our training strategy that considers the context of previously generated lines.




\vspace{-0.08in}

%% file: content/06-related_works.tex
\section{Related work}
\paragraph{Controllable lyrics generation}
While various lyric generators and AI-assisted lyrics writing system have been developed to follow control signals like themes, rhyme, specific text format, or in-filling text template \citep{Ma2020Rigid,Li2020Ai-lyricist,Liu2020ChipSong,Zhang2020Youling,sun2022songrewriter}, none provide \textit{full control} over the sentence-level semantics of generated content and follow melody constraint; \citet{Fan2019hierarchical} employed control mechanisms to generate lyrics based on specific topics, \citet{tian2023unsupervised} used keywords and genre to control the content, and \citet{Saeed2019Creativegans}, used music audio to condition the generation process. Other approaches have incorporated stylistic elements, such as rhyme schemes and meter or text format to influence the lyrical output \citep{Potash2015Ghostwriter,Zhang2020Youling}. Despite these efforts, they have insufficient sentence-level semantics and note-level music alignment. \ModelName{} can generate coherent, full-fledged high quality melody-aligned lyrics with sentence-level semantics control.
\vspace{-1mm}
\paragraph{Melody-Lyrics alignment}
LLMs have proven effective in the M2L generation task, with various attempts to integrate music representation \citep{ Lee2019icomposer, Qian2022Training}. For example, \citet{sheng2021songmass}  applied two transformers for cross-attention between lyrics and melody; \citet{tian2023unsupervised,qian2023unify} considered duration of musical note and stressed syllables mapping, beat, or song structures during lyrics generation. ReLyMe \citep{zhang2022relyme}, a lyrics-to-melody (L2M) generation model, considers the mapping of keywords to stressed beat locations but does not discuss how pitch influences the determination of prominent notes or how keywords are identified. While revision approaches have been explored in poetry generation \citep{Zugarini2021Generate}, REFFLY is the first framework for melody-constrained lyrics revision with sentence-level semantic control. Furthermore, existing M2L models fail to align prominent words with prominent notes, resulting in poor prosody. We propose a novel heuristic for melody-lyrics alignment, achieving a 26\% improvement. \ModelName{} enhances lyric generation and emotional expression. 



%% file: content/07-conclusion.tex
\vspace{-2mm}
\section{Conclusion}
\vspace{-2mm}
We introduced \ModelName{}, the first melody constrained revision framework to generate high-quality lyrics from plain text drafts while retaining original meaning. To enhance the lyrics-melody alignment, we designed a heuristic to identify and align prominent notes and words. Finally, we show \ModelName{} excel across diverse applicability.

\section*{Limitation}

The limitation of our work include: 1) In this work, we use a rule-based method to identify important words in lyrics, specifically, nouns, verbs, and adjectives. Future work could investigate more nuanced definitions of important words. 2) Similarly, Our method for extracting prominent notes considers only two levels: prominent notes and other notes. While this simple approach has yielded satisfying results, exploring more fine-grained categories could potentially enhance performance. 3) Although our approach preserves the original meaning of the lyrics, the genre of the resulting lyrics largely depends on the training dataset. Future research could aim to provide more control over the genre.

%% file: content/Appendix.tex
\clearpage

\section{Exemplary Data}

\subsection{Example of full-length Song}
\begin{figure}[!ht]
    \centering
    \includegraphics[width=1\linewidth]{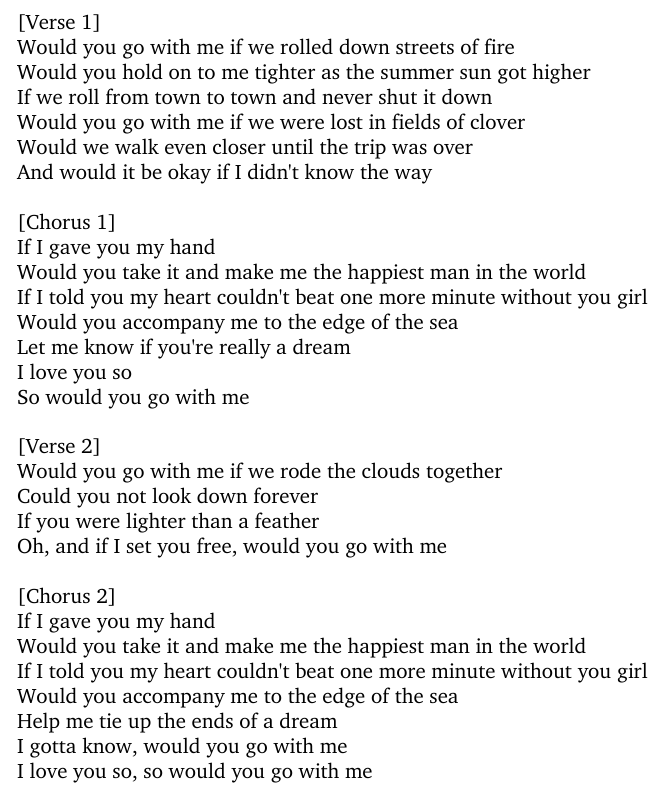}
    \caption{Example song with verse-chorus-verse-chorus structure}
    \label{fig:song_strucutre}
\end{figure}

\subsection{Example of validation dataset}
\begin{figure}[!ht]
    \centering
    \includegraphics[width=1\linewidth]{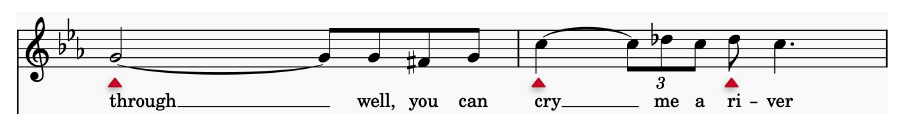}
    \caption{Exemplary data points in validation datasets, where experts annotate the ground truth prominent notes, actual data points have at three to five musical phrases.}
    \label{fig:val_datapoint}
\end{figure}
\begin{figure}[!ht]
    \centering
    \includegraphics[width=1\linewidth]{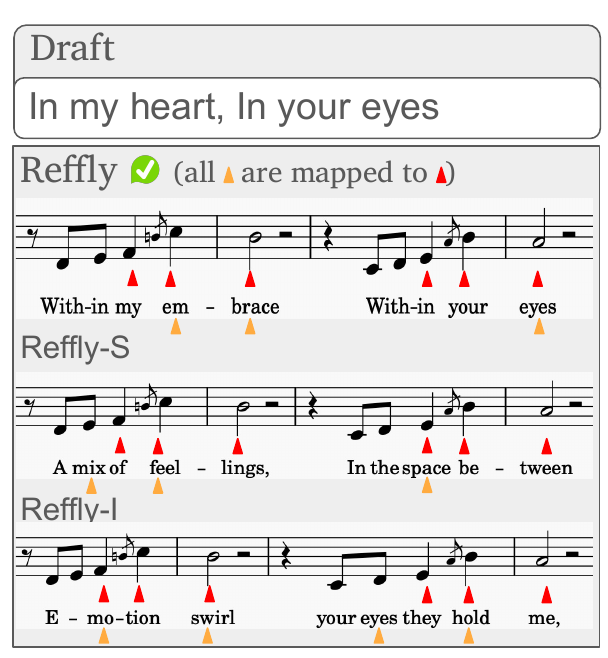}
    \caption{Example song}
    \label{fig:example_ablation}
\end{figure}

\section{Human Evaluation Details}
Human annotators are paid with \$ 20 per hours.Note that the use of an external voice synthesizer with limited quality impacts the average score of the human evaluation for the Arbitrary Generation task. Despite this, our method outperforms all baselines, with an inter-annotator agreement of 0.69, further demonstrating the effectiveness of our \ModelName{}. Given the current limitations of open-source AI singing voice generation models, the quality of the singing in the generated songs may not meet human standards, which can lead to lower scores.
It is important to note that, under the same test settings, our model performs significantly better than all the baselines in every task setting.

\subsection{Qualification Task}
\label{app:qual}
To evaluate the Turkers' expertise in the field, we designed a task that included the initial verse from 9 different songs, each with ground-truth labels. These songs were chosen with care to avoid unclear cases, allowing for a clear assessment of quality. The selected songs were those whose scores showed a strong correlation with the ground-truth labels. We select 49 qualified annotators out of 87 annotators, based on Pearson correlation metric. The average inter-rater agreement interms of Pearson correlation among qualified annotators in qualification task was 0.43. 

\subsection{Annotation Task}
We present the original survey, including evaluation instructions and the annotation task, in Figure~\ref{fig:evaluation_task_instruction} through Figure~\ref{fig:evaluation_annotation}. Figure~\ref{fig:evaluation_task_instruction}, Figure~\ref{fig:evaluation_task_instruction_examples_1}, and Figure~\ref{fig:evaluation_task_instruction_examples_2} outline task instructions, defining each metric—intelligibility, singability, prosody, coherence, creativity—and accompanied by examples of good and bad lyrics for each criterion. Figure~\ref{fig:evaluation_annotation} display the actual annotation task.
\section{Experiments Details}

\subsection{Details regarding to baseline construction}
\label{sec:baselines_detail}
\paragraph{ChatGPT} We used ChatGPT-4-turbo as the base model to construct this baseline. In order to make this baseline to be fair, we tried our best to prompt ChatGPT-4. Firstly, we use 2-shot manner to prompt it: we provide two golden exemplary revision example every time. To make sure that ChatGPT have the same information that \ModelName{} has, we provided lyrics and the corresponding serialized score using music21, a format that zero-shot ChatGPT could understand. This score encompasses  every detail about the music, including rhythm, pitch, and time signature. Note that extracting additional details, such as the position of prominent notes, would require the prominent note extractor from Reffly’s framework. Our objective is to use ChatGPT-4 as a baseline, not to replace Llama2-13b as a revision module.

\paragraph{Lyra}
Since the original Lyra paper \citep{tian2023unsupervised} used GPT-2 as the base model, in order to make the comparison fair, we re-implemented the Lyra using Llama-2-13b. When doing the experiments, we use the exact same lyrics drafts as \ModelName{}, which is generated from a collected user prompt. Since Lyra requires keywords as inputs to generate each sentence, we use Yake \citep{Ricardo2020Yake} to extract three keywords from the lyrics draft, as the same setting as the original paper.

\subsection{Example of the interface used to collect scattered user input}

\begin{figure}[h]
    \centering
    \includegraphics[width=\linewidth]{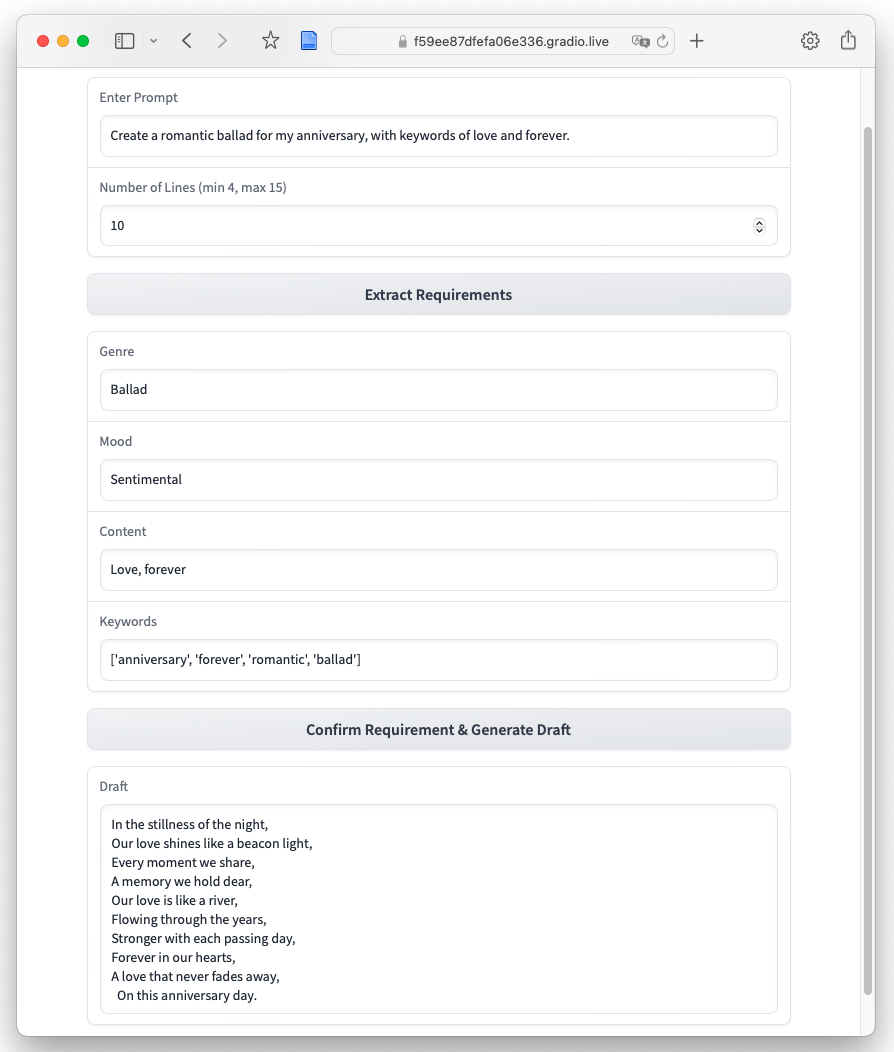}
    \caption{Interface used for input to draft}
    \label{fig:i2d}
\end{figure}

The Figure \ref{fig:i2d} illustrates the interface of our input-to-draft model. Initially, the user's requirements are extracted from the prompt using few-shot LLaMA2-13b with intent extraction examples. The extracted requirements are then presented to the user for confirmation, after which a draft is generated based on the confirmed requirements using LLaMA2-13b. 

Note that we use the same revision model in all of arbitrary generation, full-length generation, and song translation. Only the input lyrics draft is different, which are generated by LLaMA2-13b in few-shot manner.

\label{subsec:input-to-draft}

\subsection{Computation cost}

As illustrated in Figure \ref{fig:overview}, we ``restart'' if the all of the outcome of Step 3.1 is \textit{unsigable}. Restart happens when the length of lyrics draft is too different from the length of music constraint. When rephrasing, we rephrase the original draft so that the length of draft is closer to the length of music constraint. When doing the experiment, all of the lyrics are generated within 3 iterations.

\subsection{More details regarding to experiment metrics}
 We utilized the GPT-2 Large model, which has 774 million parameters, to calculate perplexity. Prominent word-note matching rate, or matching rate in Table \ref{tbl:evaluation_result}, is the accuracy of stressed syllables of prominent words being correctly mapped to prominent notes.
\section{\ModelName{} Details}

\subsection{candidate selection algorithm}
The candidate selection algorithm refines a list of lyric candidates to select the best match based on melody constraints, calculating ties and similarities to ensure a singable output. Note that there is no need for human to post process the lyrics. 

\begin{figure}[h]
    \centering
    \includegraphics[width=\linewidth]{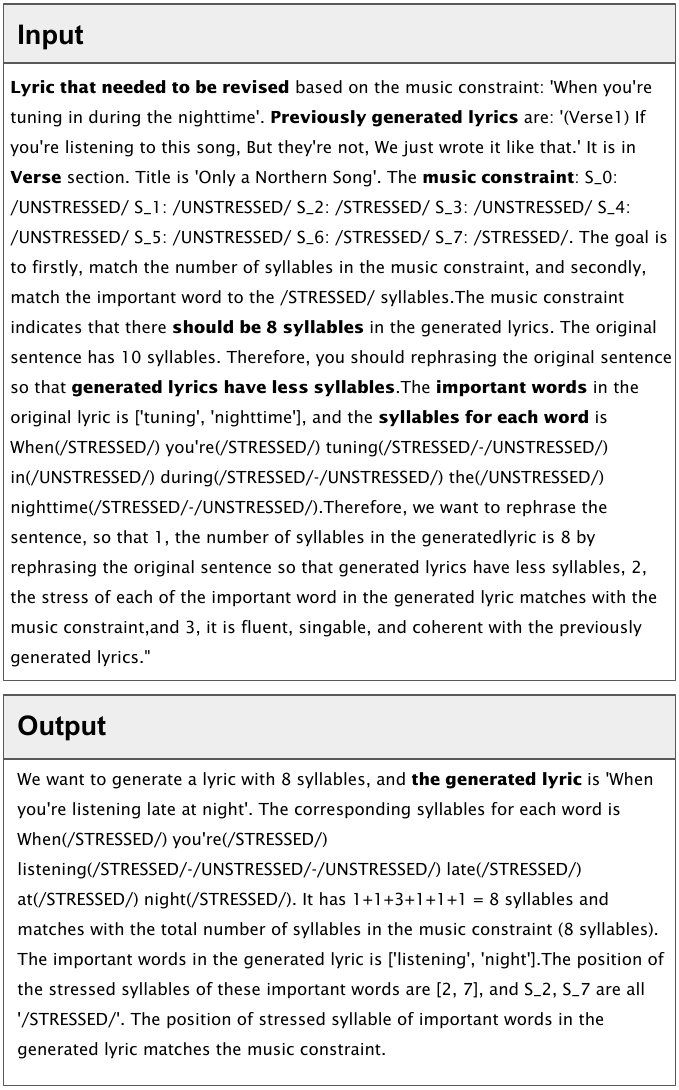}
    \caption{Training data example.}
    \label{fig:example_training_detail}
\end{figure}
\subsection{Training Data Construction}
\label{app:training_data}

\paragraph{Construct Input Data} We use ChatGPT 3.5 to rephrase sentences either trivially or non-trivially (50\% vs. 50\%). Trivial rephrasing changes only a few words without altering the sentence structure. The \textbf{rephrased lyric} is then used as the input for revision module training. Additionally, we provide \textbf{instructions} to better guide LLMs revise lyric that better align with melody. The instructions contain information about 1) the phoneme of each word sourced from the CMU pronunciation dictionary, 2) the number of syllables in both the original and rephrased sentences, and 3) guidelines for modifying the lyric drafts, such as making the sentence shorter or longer.

\paragraph{Generate Pseudo Music Constraint} The final piece of input data is \textbf{music constraint}. Due to the lack of melody-aligned data, we generate a pseudo music constraint for lyric dataset, independent of its melody. This approach addresses both the scarcity and copyright issues associated with aligned data. We assume that the lyrics in the training data exhibit good prosody and are singable. we assign a special token, symbolizing a pseudo note, to each syllable in the sentence (singability assumption). If the syllable associated with the special token occupies the stress position of a prominent word, the token denotes an prominent note, otherwise it represents a less prominent note (good-prosody assumption). This way, we ``back-translating" (\cite{Ou2023Songs}) the pseudo music constraint from pure lyric-side data.

\paragraph{Assemble Output Data} The \textbf{original lyric}, prior to rephrasing in the first step, is used as the output for revision module training. Additionally, we craft an "\textbf{explanation}" paragraph help LLMs revise the lyric by breaking down the revision task into multiple simpler sub-tasks ( more detail see Figure \ref{fig:traing}).

After processing, the inputs consist of the rephrased sentence, the song title, pseudo music constraints, the original lyrics, the song structure, and specific instructions. The output includes the original lyrics accompanied by an explanation.

We used the default LoRA implementation from the official LLaMA2-13b GitHub repository to fine-tune the model. The revision module was trained for 3 epochs using a dataset of 3,500 data points.

\paragraph{Exemplary training data point}

Figure~\ref{fig:example_training_detail} shows an example training data point constructed using the pipeline introduced in Figure~\ref{fig:traing}. The original lyric is "When you're listening late at night." We generate a "pseudo melody constraint" based on this lyric, then use ChatGPT to create a rephrased lyric draft, "When you're tuning in during the nighttime." The model is trained to generate the correct original lyric using the title, lyric draft, pseudo melody constraint, and instruction as input.

\section{Music Theory}
\subsection{Representation in Melody}
The representation for a melody is hierarchical. A melody M consists of a series of musical phrase $M = (p_0, p_2, .. p_x)$, where x is the total number of musical phrase; Each musical phrase consists of a series of measures $p_i = (m_1, m_2,...m_y), i\in [0,x]$, where y is the total number of measures in i'th musical phrase. Note that $|p_j \cap p_{j+1}| \leq  1$. The intersection equals to 1 when a musical phrase end in the middle of a certain measure, so the next musical phrase starts from the same measure. Each measure consists of a series of notes and a corresponding time signature. $m_k = (n_1,n_2,...,n_z)$, where z is the total number of notes in measure $m_k$. Each note has four component: pitch, duration, offset, and tie. Pitch represents the highness/lowness of a note; duration is the length of the note; offset is the beat when this note starts in its measure; and tie can be start (a tie starts from this note), or continue (in between of a tie), or end (a tie ends at this note).

\subsection{Prominent note extraction heuristic details }
\label{subsec:prominent_note_details}
Inspired by prior research in music theory \cite{Palmere4cfret2006what}, we develop  a more comprehensive heuristics to identify prominent musical notes based on three fundamental characteristics of music:
\begin{enumerate}
\item \textit{Time Signature}: This characteristic provides a structured framework that dictates how beats are grouped and accented within each measure. We identify notes that fall on strong beats or downbeats as prominent notes.
\item \textit{Rhythm}: For this characteristic, we specifically examine \textit{syncopation}, a musical technique that shifts emphasis to beats or parts of a beat where it is not usually expected. This technique breaks the conventional rhythmic pattern by highlighting off-beats or weaker beats within a measure. Notes that are accentuated using this technique are identified as prominent notes.
\item \textit{Pitch}: From this characteristic, we particularly focus on \textit{pitch jump}. Large pitch jumps contribute to contrast and variety in the melody line, thereby making notes with significant pitch jumps more conspicuous. We classify notes that exhibit significant pitch jumps as prominent notes.
\end{enumerate}
\textbf{Melody}
Melody is a sequence of musical tones, consisting of multiple musical phrases that can be further decomposed into timed musical notes. Each musical note has two independent aspect: pitch and duration. Pitch refers to the perceived highness or lowness of a sound; duration refers to the length of time that a musical note is held or sustained. \\
\textbf{Time signature}
time signature organizes the rhythm and provides a framework for how the beats are grouped and accented within each measure. 
A time signature is represented by two numbers, one stacked on top of the other: the top number indicates the number of beats in each measure; the bottom number indicates the duration value that represents one beat. For example, 4/4 means a quarter note as one beat, 4 beats in a measure. Table 1 shows the stressed location for some commonly-seen time signature. The elements in the list is the number of beat that is stressed. For example, [0,2] means the first and third beats are stressed.
\begin{table}[h] 
\centering
\caption{Time signatures and their stressed locations}
\begin{tabular}{cc}
\toprule
\textbf{Time Signature} & \textbf{Stressed Location} \\
\midrule
4/4 & [0, 2] \\
3/4 & [0] \\
2/4 & [0] \\
3/8 & [0, 2] \\
6/8 & [0, 2] \\
9/8 & [0, 2, 5] \\
12/8 & [0, 2, 5, 8] \\
\bottomrule
\end{tabular}
\label{tab:time_sig}
\end{table}\\
\textbf{Syncopation}
Syncopation refers to the displacement or shifting of accents or emphasis to unstressed beats. If a note is in unstressed beat with a longer duration than its previous note, then this note, although in unstressed beat, is stressed, or syncopated. \\
\textbf{Pitch jump}
Pitch jump for two consecutive notes is the absolute difference of their pitch value. Larger pitch jumps create contrast and variety within the melody line. 

If a note is in a metrical stressed position (indicated by time signature), or it is a syncopation, or it has a pitch jump (larger than average interval), we consider it as a prominent note, otherwise, it is a non-prominent note.\\

We also provide the mathmatical formulations for prominent note extraction:\\
\textbf{1. Time signature:}\\
As shown in table \ref{tab:time_sig}, a function can determine if a note is in an important location in terms of time signature. Suppose the time signature for this melody is $T$, and the corresponding list for stressed location is $SL_T$
\begin{equation}
\text{stressed}(n_i) = 
\begin{cases} 
1 & \text{if }\text{if offset}(n_i) \text{ in }SL_T \\
0 & \text{, otherwise}
\end{cases}
\end{equation}
\textbf{2. Rhythmic type: }

We implemented two simple rules:

1) for a given k consecutive notes $n_i,...n_{i+k}$ that are connected by one tie, we replace $n_i,...n_{i+k}$ as a new note $n_i'$, and the duration for $n_i'$ is\\ $duration(n_i') = \sum_{j = 0}^{k} duration(m_{i+j})$, \\and offset (beginning of $n_i'$) is:\\$offset (n_i') = offset (n_i)$

2) After combining all notes that connected by tie, we check syncopation. If a note is in a weak location but its duration is longer than previous note, it is a syncopation.

\noindent \textbf{3. Pitch:}\\
A note is more important if there as a dramatic change in pitch compared to previous note. Based on this assumption, we have
\begin{equation}
\text{jump}(n_i) = 
\begin{cases}
1 & \text{if } \Delta\text{pitch}(n_i) > AIJ, \\
0 & \text{otherwise},
\end{cases}
\end{equation}
where 
\begin{align*}
\Delta\text{pitch}(n_i) &= | \text{pitch}(n_i) - \text{pitch}(n_{i-1})|, \\
AIJ &= \frac{\sum_{i=1}^{x} \Delta\text{pitch}(n_i)}{x}, \\
\text{and } x &\text{ is the number of notes.}
\end{align*}

\textbf{Collectively}, the importance of note $m_i$ is defined by\\
\begin{equation}
\text{M}(m_i) = 
\begin{cases} 
1 & \text{if }\text{if stressed, jump, or syncopate}(m_i) \\
0 & \text{, otherwise}
\end{cases}
\label{m_n}
\end{equation}
, where (0)1 means the note is an (un)important note.

\subsection{Results for heuristics}
\label{subsec:Results_heuristics_details}
We provide more details for $\S$ \ref{sec:effectiveness_heuristics} at here. Because our validation dataset only contains ground truth prominent note, we use Yake \citep{Ricardo2020Yake} algorithm to extract up to 3 keywords from one lyrics sentence, and treat the extracted keywords that correspond to ground truth prominent notes as ground truth prominent words. \\
Note that in the heuristic, we are consider the stressed syllable position by aligning stressed syllables of prominent words to prominent notes. 
\begin{figure*}[h]
  \centering
  \includegraphics[width=\textwidth]{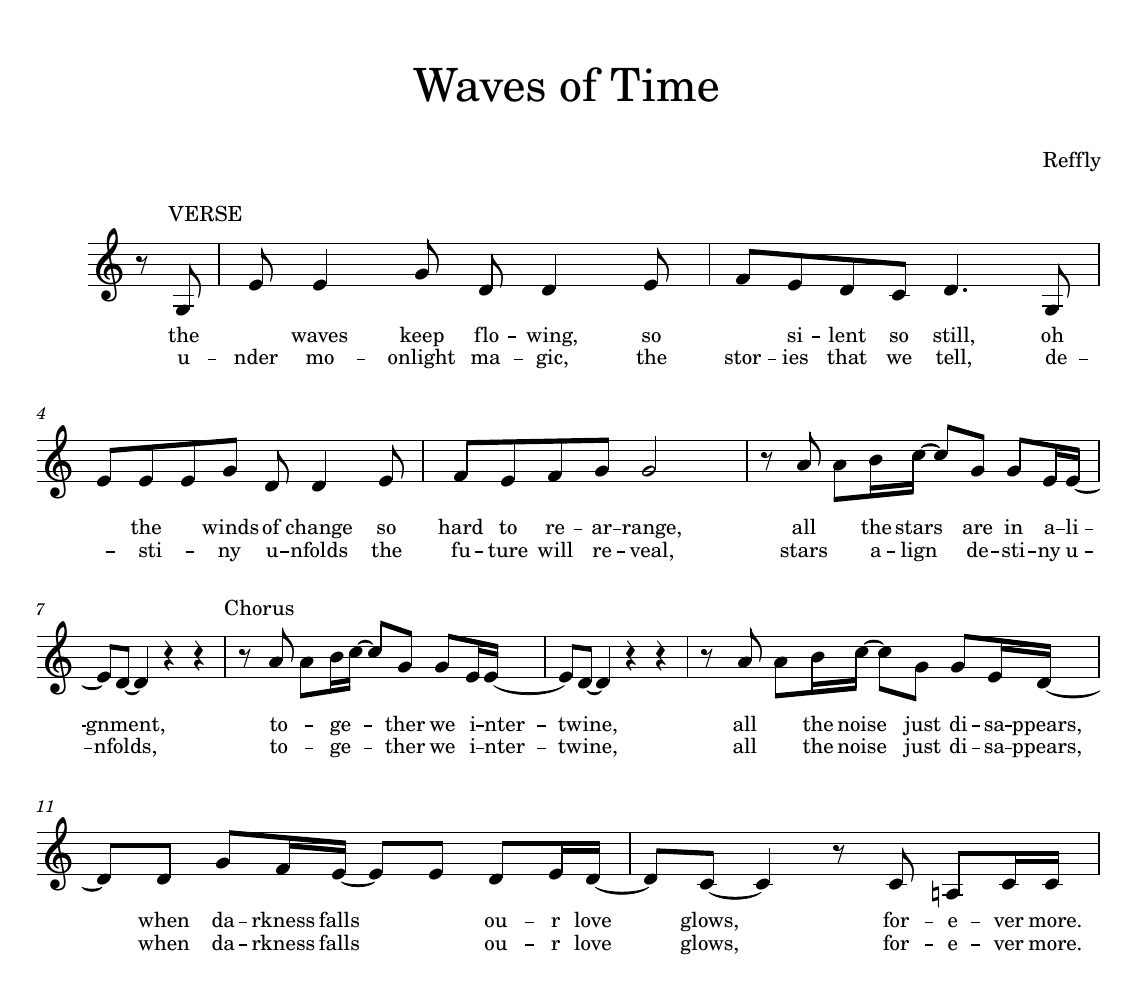}
  \caption{Exemplary generated song with verse-chorus-verse-chorus structure } 
  \label{fig:example_struct_output}
  \end{figure*}

\begin{figure*}[h]
  \centering
  \includegraphics[width=\textwidth]{figures/Appendix/evaluation_task_instruction.pdf}
  \caption{Human evaluation survey: task instruction} 
  \label{fig:evaluation_task_instruction}
  \end{figure*}
  \begin{figure*}[h]
  \centering
  \includegraphics[width=\textwidth]{figures/Appendix/evaluation_task_instruction_examples_1.pdf}
  \caption{Human evaluation survey: explanation of different metrics 1} 
  \label{fig:evaluation_task_instruction_examples_1}
  \end{figure*}
  \begin{figure*}[h]
  \centering
  \includegraphics[width=\textwidth]{figures/Appendix/evaluation_task_instruction_examples_2.pdf}
  \caption{Human evaluation survey: explanation of different metrics 2} 
  \label{fig:evaluation_task_instruction_examples_2}
  \end{figure*}
  \begin{figure*}[h]
  \centering
  \includegraphics[width=\textwidth]{figures/Appendix/evaluation_annotation.pdf}
  \caption{Human evaluation survey: the annotation task} 
  \label{fig:evaluation_annotation}
  \end{figure*}
\subsection{An illustrative example of how the tie is added}
 In figure \ref{fig:example_case_study}, we need to add two ties to the first sentence ``eyes of peace a canvas of hues'', because there are two more notes in the melody than number of syllables. Here, we discuss why there is a tie added to ``Do'' instead of ``Mi''. This is because at here, ``Mi'' would be identified as a prominent note (because it is in a stressed beat in 4/4) and ``peace'' as a prominent word (because it is a noun). Note that here ``Do'' is not a prominent note, because it is neither in a stressed location, nor a syncopated note, nor a note with big pitch jump from the previous note. To maximize the number of prominent words mapped to prominent notes, a tie would be added at ``Do'', and the important word ``peace'' is corresponding to the prominent note ``Mi''. This process is detailed at Step 3.2 ($\S$\ref{sec:alignement}) and Algorithm \ref{algo:alignment}, candidate selection algorithm. This entire process is handled algorithmically, without requiring human inspection.